\pdfoutput=1
\documentclass[11pt]{article}

\usepackage{emnlp2021}
\usepackage{booktabs} % To thicken table lines

\usepackage{times}
\usepackage{latexsym}

\usepackage[T1]{fontenc}
\usepackage{amsmath}
\usepackage[utf8]{inputenc}

\usepackage{microtype}
\usepackage{graphicx,subcaption}

\usepackage[ruled,vlined]{algorithm2e}
\usepackage{algpseudocode}
\SetKwInput{kwInput}{Input}
\SetKwInput{kwOutput}{Output}
\SetKw{Break}{break}
\usepackage{float}
\newcommand{\red}[1]{\textcolor{red}{#1}}

\title{
    Adversarially Constructed Evaluation Sets Are More Challenging, \\but May Not Be Fair
}

\author{
  Jason Phang,$^1$
  Angelica Chen,$^1$
  William Huang,$^2$\footnotemark 
  \ \ Samuel R. Bowman${}^1{}^3{}^4$\\
\textnormal{$^1$Center for Data Science, New York University} \\
\textnormal{$^2$Capital One} \\
\textnormal{$^3$Dept. of Linguistics, New York University} \\
\textnormal{$^4$Dept. of Computer Science, New York University}
\AND \textbf{Correspondence}: \href{mailto:jasonphang@nyu.edu}{\tt jasonphang@nyu.edu}
}

\begin{document}
\maketitle
\begin{abstract}
  More capable language models increasingly saturate existing task benchmarks, in some cases outperforming humans.
  This has left little headroom with which to measure further progress.
  Adversarial dataset creation has been proposed as a strategy to construct more challenging datasets, and two common approaches are: (1) filtering out easy examples and (2) model-in-the-loop data collection.
  In this work, we study the impact of applying each approach to create more challenging evaluation datasets.
  We adapt the AFLite algorithm to filter evaluation data, and run experiments against 18 different adversary models.
  We find that AFLite indeed selects more challenging examples, lowering the performance of evaluated models more as stronger adversary models are used.
  However, the resulting ranking of models can also be unstable and highly sensitive to the choice of adversary model used.
  Moreover, AFLite oversamples examples with low annotator agreement, meaning that model comparisons hinge on the most contentiously labeled examples.
  Smaller-scale experiments on the adversarially \textit{collected} datasets ANLI and AdversarialQA show similar findings, broadly lowering performance with stronger adversaries while disproportionately affecting the adversary model.
\end{abstract}
\renewcommand*{\thefootnote}{\fnsymbol{footnote}}
\footnotetext[1]{Work done while at NYU.}
\renewcommand*{\thefootnote}{\arabic{footnote}}
\setcounter{footnote}{0}

\section{Introduction}

Large-scale language models have attained strong performance across a variety of language understanding tasks, including question-answering, natural language inference (NLI), and paraphrase identification.
As the capabilities of these models improve, it has become increasingly difficult to systematically evaluate and benchmark further model improvements \citep{vania-etal-2021-comparing}.
Standard benchmarking tasks such as SQuAD \citep{rajpurkur2016squad, lee-etal-2020-squad2} and multi-task benchmarks such as GLUE \citep{wang-etal-2018-glue} and SuperGLUE \citep{wang2019superglue} have seen models attain scores higher than human baseline scores. 
This has left little headroom with which to measure further improvements in models and progress in NLP.
More than ever, we need new approaches to build challenging and reliable evaluation datasets at scale \citep{bowman2021fix}.

Prior work such as \citet{bras2020aflite} and \citet{nie-etal-2020-adversarial} have proposed adversarially filtering or constructing datasets to raise the difficulty of natural language understanding tasks, by leveraging existing highly capable models to assist with test example selection or creation.
However, one potential issue is that an adversarially constructed dataset that targets a specific model may bias the resulting data, creating datasets that may be unduly challenging for one class of models but not others. 
In the extreme, adversarial datasets may be so narrowly optimized toward stumping a particular model that they no longer accurately measure the abilities that the dataset was designed to test.

In contrast to prior work focusing on adversarial dataset creation for training \citep{wallace2021dadc} or training and evaluation data \citep{bras2020aflite, nie2020anli}, we focus solely on evaluation data, and whether the choice of the adversary model can introduce unwanted biases into an evaluation dataset. 
% For instance, one could imagine an adversarially created dataset that isolates a particular linguistic phenomenon which the adversary model has learned incorrectly, resulting in a dataset on which the adversary model (and closely related models) performs extremely poorly but that is no more challenging for other models.
Ideally, an adversarially created dataset should be more difficult for all models, regardless of the choice of the adversary.

In this work, we investigate two different approaches to create a more challenging task evaluation dataset using adversary models: adversarial filtering, which filters out examples from a static dataset that are identified to be easy for a given adversary model, and model-in-the-loop adversarial data collection, where human annotators are tasked with interactively creating examples that stump an adversary model.

For adversarial filtering, we study AFLite \citep{sakaguchi2020winogrande, bras2020aflite}, an algorithm that identifies challenging subsets of a given task dataset. We apply AFLite in extensive experiments across four commonly used English-language NLP datasets and 18 different models to study the interaction between the choice of adversary model and the resulting evaluation performance.
For adversarial data collection, we evaluate a range of models against two existing adversarially generated datasets using a model in the loop: ANLI \citep{nie-etal-2020-adversarial} and AdversarialQA \citep{bartolo2020beat}.

We find that adversarial filtering and adversarial dataset collection do result in more challenging evaluation datasets, but they are not without their drawbacks. 
We find that the general outcome of adversarial filtering is to lower performance across the board, with stronger adversary models leading to more challenging subsets of examples. 
However, as more difficult evaluation subsets are identified, the relative order of model performance is not preserved, with large random variation in model ranks as stronger adversaries are used.
This suggests that using adversarially filtered datasets for benchmarking models can be potentially problematic.
The performance on the resulting datasets is also much worse if the evaluated model is fine-tuned from the same base model as the adversary.
In other words, the difficulty of an adversarially filtered task may be overstated if evaluated on the same pretrained model.
We also show that adversarial filtering tends to oversample examples with low annotator agreement, which could mean that these examples are contentious even for human annotators.
While \citet{pavlick-kwiatkowski-2019-inherent} and \citet{nie-etal-2020-learn} show there is often genuine disagreement over the labels of the more challenging NLI examples and that low agreement does not entail label noise, they also argue that such evaluation on these examples need to take into account the label disagreement, and simply computing accuracy can be misleading.
% These issues, while not fatal, suggest that using adversarially filtered datasets for benchmarking models can be potentially problematic.

Similarly, we find that the adversarially \textit{collected} datasets ANLI \citep{nie-etal-2020-adversarial} and AdversarialQA \citep{bartolo2020beat} are also more challenging for all models while also showing signs of disproportionately disadvantaging the adversary model.
However, with only a small number of adversarial datasets available, it is more difficult to draw strong conclusions about the overall efficacy or potential drawbacks of the approach.

In both cases, our findings do not preclude the viability of adversarial dataset creation for evaluation purposes, but we urge researchers to keep these issues in mind when evaluating or comparing models based on adversarial datasets.

\section{Related Work}
  
The AFLite adversarial filtering algorithm that we perform most of our experiments on was first proposed by \citet{sakaguchi2020winogrande}. The same work also introduced Winogrande, an adversarial Winograd Schema Challange dataset. \citet{bras2020aflite} later provided further theoretical and empirical justification for the algorithm, showing that models train on AFLite-filtered data generalize better to out-of-domain datasets.

Other datasets have used variants of adversarial filtering, using a model to filter out easy examples from a dataset. \citet{zellers2018swagaf} introduced SWAG, an adversarially filtered commonsense multiple-choice dataset, and \citet{zellers2019hellaswag} introduced HellaSwag, a follow-up using better-performing adversary models. Both datasets also used additional text generation models to create incorrect options.

An alternative approach to model-adversarial dataset creation is to collect data using a model in the loop, where during the process of writing examples, human crowdworkers are given immediate feedback on whether a trained adversary model is able to correctly answer their example, and are incentivized to write examples on which the models fail. \citet{nie2020anli} introduce ANLI, an adversarial natural language inference dataset, using BERT and RoBERTa as adversary models. \citet{williams2020anlizing} provide fine-grained analysis of the kinds of examples arising from this adversarial dataset creation procedure. \citet{bartolo2020beat} introduce AdversarialQA, a question-answering dataset using a trained BiDAF, BERT and RoBERTa models as adversaries. \citet{kiela2021dynabench} further extend this approach, building a platform for continuous human-and-model-in-the-loop data creation. Using adversarially collected data as training data has been shown to lead to better performance on other adversarial datasets, but worse on out-of-domain datasets \citep{kaushik-etal-2021-efficacy,bowman-etal-2020-new}. However, models trained on adversarially collected data through many successive rounds have been shown to attain better performance \citep{wallace2021dadc}.

% \citep{mccoy-etal-2020-berts} Generalization out of domain can vary wildly across different random restarts

\section{Adversarially Filtering Evaluation Sets}

\begin{algorithm}[t]
\SetInd{0.4em}{0.7em}
\footnotesize
\SetAlgoNoEnd
\kwInput{training dataset $D_T=(X_T,Y_T)$, \red{evaluation dataset $D_V=(X_V,Y_V)$}, pre-computed representation $(\Phi(X_T)$,\red{$\Phi(X_V)$}$)$, model family $\mathcal{M}$, target dataset size $n$, number of random partitions $m$, training set size $t<n$, slice size $k\leq n$, early-stopping threshold $\tau$}
\kwOutput{\red{Filtering history of evaluation examples $H$, remaining evaluation examples $R$}}
 $S=D_T$ \\
 \red{$R=D_V$} \\
 \While{$|S| > n$}{
  \texttt{// Filtering phase} \\
  \ForAll{$i\in S$}{
    Initialize multiset of out-of-sample training predictions $E_T(i)$\;
  }
  \red{\ForAll{$i\in R$}{
    Initialize multiset of out-of-sample evaluation predictions $E_V(i)$\;
  }}
  \For{\textup{iteration} j : 1..m}{
    Randomly partition $S$ into $(T_j, S\setminus T_j)$ s.t. $|S\setminus T_j|=t$\;
    Train a classifier $\mathcal{L}\in \mathcal{M}$ on $\{(\Phi(x), y)|(x,y) \in S\setminus T_j\}$\;
    \ForAll{$i=(x,y) \in T_j$}{
      Add the prediction $\mathcal{L}(\Phi(x))$ to $E_T(i)$\;
    }
    \red{\ForAll{$i=(x,y) \in R$}{
      Add the prediction $\mathcal{L}(\Phi(x))$ to $E_V(i)$\;
    }}
  }
  \ForAll{$i=(x,y) \in S$}{
    Compute the predictability score $\tilde{p}(i)=|\{\hat{y} \in E_T(i)$ s.t. $\hat{y}=y\}|/|E_T(i)|$\;
  }
  \red{\ForAll{$i=(x,y) \in R$}{
    Compute the predictability score $\tilde{p}(i)=|\{\hat{y} \in E_V(i)$ s.t. $\hat{y}=y\}|/|E_V(i)|$\;
  }}
  Select up to $k$ instances $S'$ in $S$ with the highest predictability scores subject to $\tilde{p}(i) \geq \tau$\;
  $S=S \setminus S'$\;
  \red{Select all instances $R'$ in $R$ where $\tilde{p}(i) \geq \tau$}\;
  \red{$R=R \setminus R'$}\;
  \red{Append $R'$ to $H$}\;
  \If{|S'| < k}{
   \Break\;
  }
 }
 \Return \red{$H, R$}
 \caption{AFLite for Evaluation Data}
 \label{algo:aflite_modified}
\end{algorithm}

AFLite \citep{sakaguchi2020winogrande,bras2020aflite} is an adversarial filtering algorithm that involves iteratively removing ``easy'' examples from a dataset through multiple rounds of filtering. 
First, given a dataset $D=(X, Y)$ of inputs $X$ and labels $Y$, we compute a learned representation for each example $\Phi(X)$ based on the adversary model.
For instance, if the adversary is BERT, $\Phi(X)$ could be the \texttt{CLS} embeddings of BERT fine-tuned on a separate held-out training set for the task.
In each round, we sample multiple random subsets of the remaining data, fit weak classifiers on the data subsets and compute predictions on the remaining examples. 
If an example is predicted correctly by more than some threshold $\tau$ of weak classifiers, it is removed from the dataset. 
This procedure is repeated until the number of examples removed in an iteration falls below a set threshold, resulting in a reduced dataset. 
More details can be found in the original manuscript \citep{bras2020aflite}.

In contrast to \citet{sakaguchi2020winogrande} and \citet{bras2020aflite} who apply AFLite before performing a train/validation/test split, we are interested in the impact of the adversarial filtering only on evaluation datasets.
The key distinction here is that we do not want to use the examples in the evaluation datasets to train the weak classifiers or influence the filtering procedure.
Hence, we tweak the AFLite algorithm to apply filtering procedure to the evaluation examples separately.
In our experiments, we use the validation set of each task as the evaluation set.
We accomplish this by first computing embeddings for both the training and evaluation sets.
We then run the standard AFLite procedure on the training examples, but in each filtering round, we use the same weak classifiers and apply the same removal criteria to filter out ``easy'' evaluation examples. This modified procedure differs from the standard AFLite in two key ways: 
  (1) There is no limit to how many evaluation examples can be removed in each round. 
      The result of this is that it is common for many examples to be removed in the very first round of filtering.
  (2) The evaluation examples are not used in the fitting steps of the AFLite algorithm.

In Algorithm~\ref{algo:aflite_modified}, we show our modified AFlite, where the original algorithm applied to training examples is shown in black, and the additional lines applied to the evaluation examples are highlighted in red.

\section{Experimental Setup}

\paragraph{Models} The crux of our investigation is how the filtered dataset changes based on the choice of the adversary model and resulting $\Phi(X)$. 
We consider a relatively diverse set of pretrained transformer models:
    % GLoVE \citep{pennington2014glove},
    BERT \citep{devlin-etal-2019-bert},
    RoBERTa \citep{liu2019roberta}, 
    ALBERT \citep{Lan2020ALBERT},
    XLM-R \citep{conneau2020xlmr},
    ELECTRA \citep{clark2020electra},
    MiniBERTa \citep{zhang2020miniberta},
    BART \citep{lewis2020bart},
    % and GPT-2 \citep{radford2019gpt2}.
    and DeBERTa v2 and v3 \citep{he2021deberta}.
We detail the versions of each model used in Table~\ref{tab:models} in the Appendix.

% We use \textit{Base} and \textit{Large} versions of all BERT-type models as well as BART, the 1M and 1B versions of MiniBERTa, and the 117M and 345M versions of GPT-2.
% For GLoVE, we use a bag-of-words average over the per-token embeddings in the input for $\Phi$, without further tuning.
% For all remaining models, we fine-tune on a small separately held out training and validation set, following the AFLite procedure.
% We use the corresponding \texttt{CLS} or \texttt{SOS} embeddings for the BERT-type models and the last token representations for BART and GPT-2 to compute $\Phi(X)$.

\paragraph{Tasks} We consider four task datasets for our experiments.
MNLI \citep{williams2018mnli} and SNLI \citep{bowman2015snli} are natural language inference tasks, while Cosmos QA \citep{huang2019cosmosqa} and SocialiQA \citep{sap-etal-2019-social} are multiple-choice commonsense reasoning tasks. 
These tasks are chosen based on several criteria: having a large enough training set to be suitable for AFLite, being in a format suitable for AFLite (i.e. classification), and no model-adversarial procedure already having been applied in the creation of the dataset.
All four tasks are scored with simple accuracy.

\paragraph{Fine-Tuning} For all models, we execute two separate fine-tuning setups.
First, we perform full fine-tuning\footnote{Unlike in the DeBERTa paper, we do not apply SiFT during fine-tuning.} on the training set: 3 epochs for MNLI and SNLI, and 5 epochs for Cosmos QA and SocialIQA.
We repeat this across three random restarts.
Secondly, to supply the representations $\Phi(X)$ for the weak classifiers used in AFLite, we perform fine-tuning on a subset of training examples: 10\% of the training examples for MNLI and SNLI, and 5000 examples for Cosmos QA and SocialIQA, fixed across all models.
Because the AFLite procedure could be affected by the representations learned from this smaller number of examples, we also repeat this subsampling procedure across three random seeds, and perform fine-tuning and AFLite for each one.
All of our results on AFLite are averaged across these 3 random fine-tuning runs and 3 random AFLite runs.
In both fine-tuning setups, we hold out 500 examples from the training set for early stopping.
These training examples are held out for both fine-tuning as well as the AFLite procedure.
Conversely, the validation examples are only ever used in the AFLite procedure when applying filtering to evaluation examples. 

\begin{table}[ht!]
\centering
\resizebox{0.48\textwidth}{!}{\small
\begin{tabular}{lcccc}
\toprule
    Model & MNLI & SNLI & Cosmos & SIQA \\
    \midrule
    MiniBERTa-S-1M & 60.2 & 73.4 & 41.6 & 42.4 \\
    MiniBERTa-B-1B & 79.3 & 87.2 & 55.0 & 57.3 \\
    BERT-Base & 82.7 & 89.5 & 57.8 & 59.8 \\
    XLM-R-Base & 81.2 & 87.4 & 59.3 & 63.1 \\
    BART-Base & 84.6 & 89.8 & 63.4 & 65.2 \\
    BERT-Large & 85.5 & 91.0 & 61.9 & 65.5 \\
    ALBERT-Large & 86.3 & 89.9 & 62.3 & 68.5 \\
    RoBERTa-Base & 86.1 & 91.1 & 67.1 & 69.6 \\
    ALBERT-XLarge & 87.2 & 91.6 & 70.9 & 71.2 \\
    XLM-R-Large & 88.3 & 90.8 & 70.6 & 72.5 \\
    ELECTRA-Base & 87.4 & 91.5 & 69.9 & 73.4 \\
    BART-Large & 89.1 & 91.2 & 76.7 & 77.3 \\
    DeBERTa\textsubscript{RTD}-Base & 89.8 & 92.6 & 74.4 & 77.7 \\
    RoBERTa-Large & 89.6 & 91.8 & 78.5 & 77.4 \\
    ELECTRA-Large & 90.3 & 92.7 & 83.2 & 79.7 \\
    DeBERTa-Large & 90.5 & 92.7 & 85.5 & 79.1 \\
    DeBERTa-XLarge & 90.2 & 92.7 & 87.0 & 78.1 \\
    DeBERTa\textsubscript{RTD}-Large & 90.8 & 93.1 & 87.6 & 81.2 \\
\bottomrule
\end{tabular}
}
\caption{
    Performance (accuracy\%) of fully fine-tuned models on full validation sets. 
    Models are sorted in order of average performance across all four tasks.
}
\label{tab:fully_tuned}
\end{table}

\begin{figure*}[ht]
    \includegraphics[width=\linewidth]{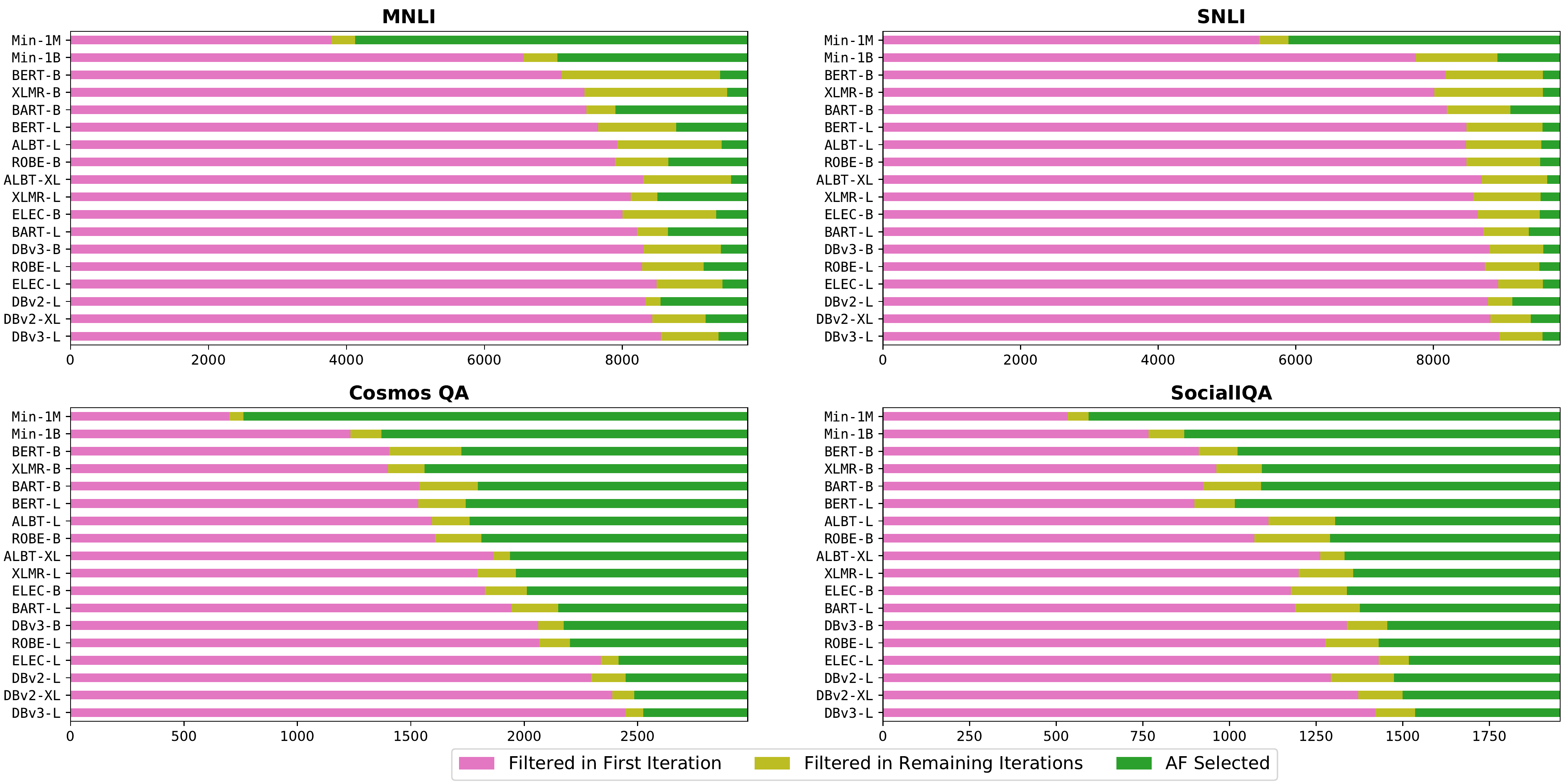}
  \caption{
    Statistics of AFLite-filtered datasets.
    We apply Algorithm~\ref{algo:aflite_modified} to the validation set of each task across adversary models, and average across three random seeds.
    \textit{AF Selected} indicates examples that are not filtered out.
    For most models, majority of the examples are filtered out within the first iteration of AFLite.
  }
  \label{filtering_statistics}
\end{figure*}

All models are trained using the \texttt{jiant} \citep{phang2020jiant} library, which is built on Transformers \citep{wolf-etal-2020-transformers} and PyTorch \citep{pytorch2019}.

We show in Table~\ref{tab:fully_tuned} the performance of our fully fine-tuned models on the validation set of each task. 
In this and subsequent visualizations, we sort the models based on the average full fine-tuned performance on the four tasks, from weakest to strongest.

\section{Adversarial Filtering of Evaluation Sets}

\subsection{AFLite Filtering Statistics}

We show in Figure~\ref{filtering_statistics} the breakdown of the result of applying AFLite with different models.
Each example in a validation set can be categorized in one of three groups: examples filtered on the first iteration of the AFLite algorithm, examples filtered in all subsequent AFLite iterations, and examples remaining after applying AFLite (AF Selected). 
Many examples, in most cases more than half the validation datasets, are filtered out within the first iteration, meaning that their labels were largely correctly predicted by a set of weak classifiers using the learned representations of partially tuned adversary models. 
Moreover, the stronger the adversary model, the more examples tend to be removed in the first iteration.
The subsequent iterations of filtering remove comparatively much fewer examples.

Among the AF Filtered examples for Cosmos QA and SocialIQA, we see a trend that the stronger the adversary model, the fewer examples remain after AFLite, meaning that fewer examples from the validation set are considered challenging.
We do not see the same pattern in MNLI and SNLI, where aside from the weakest MiniBERTa-1M model, the number of AF Filtered examples does not vary consistently across strength of models.
We note that Cosmos QA and SocialIQA use different AFLite hyperparameters from MNLI and SNLI because of the difference in datasets sizes (see Table~\ref{table:aflite_hyperparams} in the Appendix).

\begin{figure*}[ht!]
    \includegraphics[width=\linewidth]{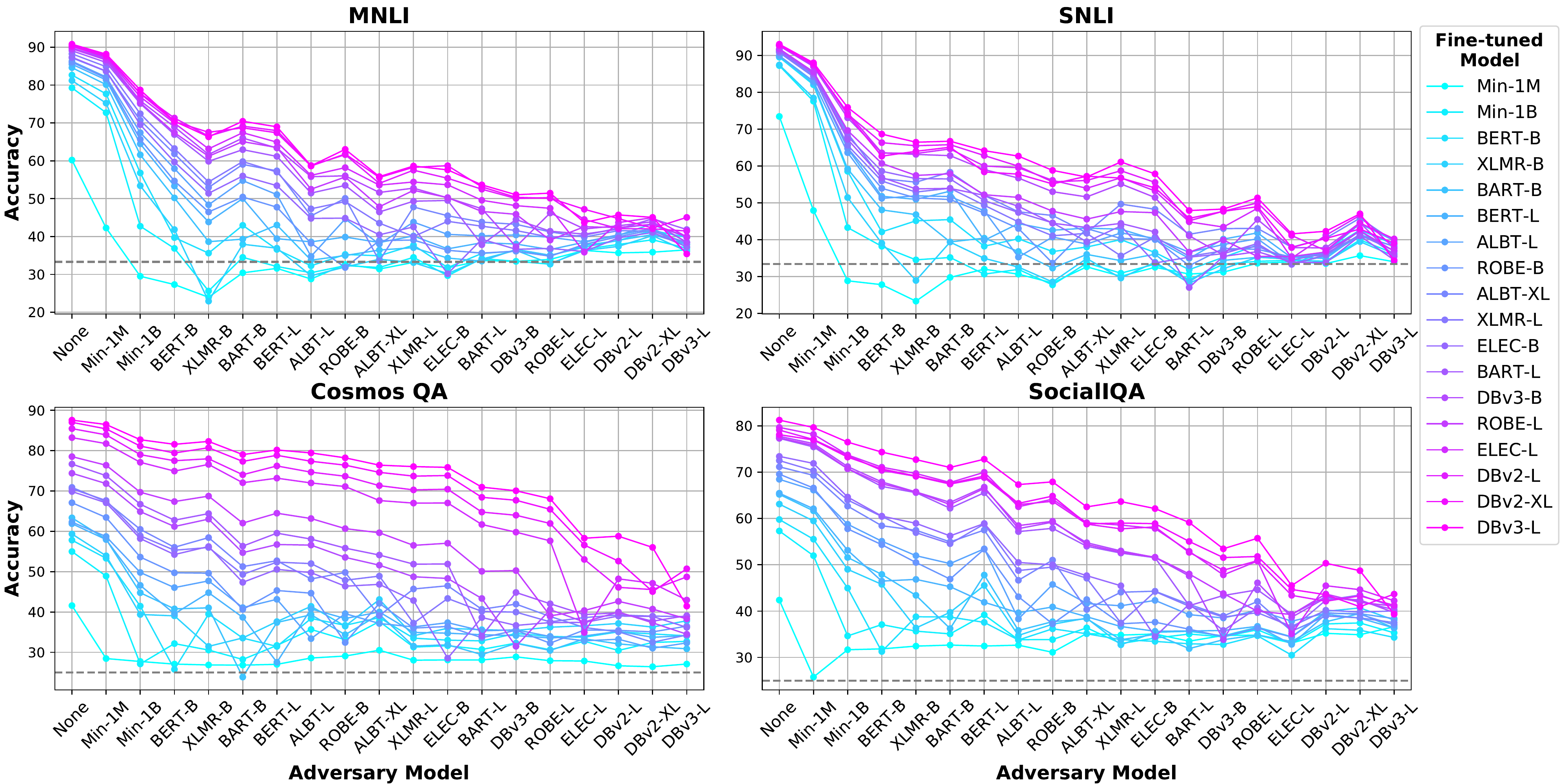}
  \caption{
    Performance of fine-tuned models on validation sets filtered via AFLite using adversary models.
    `None' indicates the full validation set with no filtering applied.
    The dotted line indicates performance at chance for each task.
    Filtering with stronger adversary models leads to lower performance on the filtered dataset, across all fine-tuned models.
    % However, filtering also tend to hurt the adversary model itself more than other models on average (darker cells on the diagonal).
  }
  \label{fig:all_scores_curves}
\end{figure*}

\begin{figure*}[ht!]
    \includegraphics[width=\linewidth]{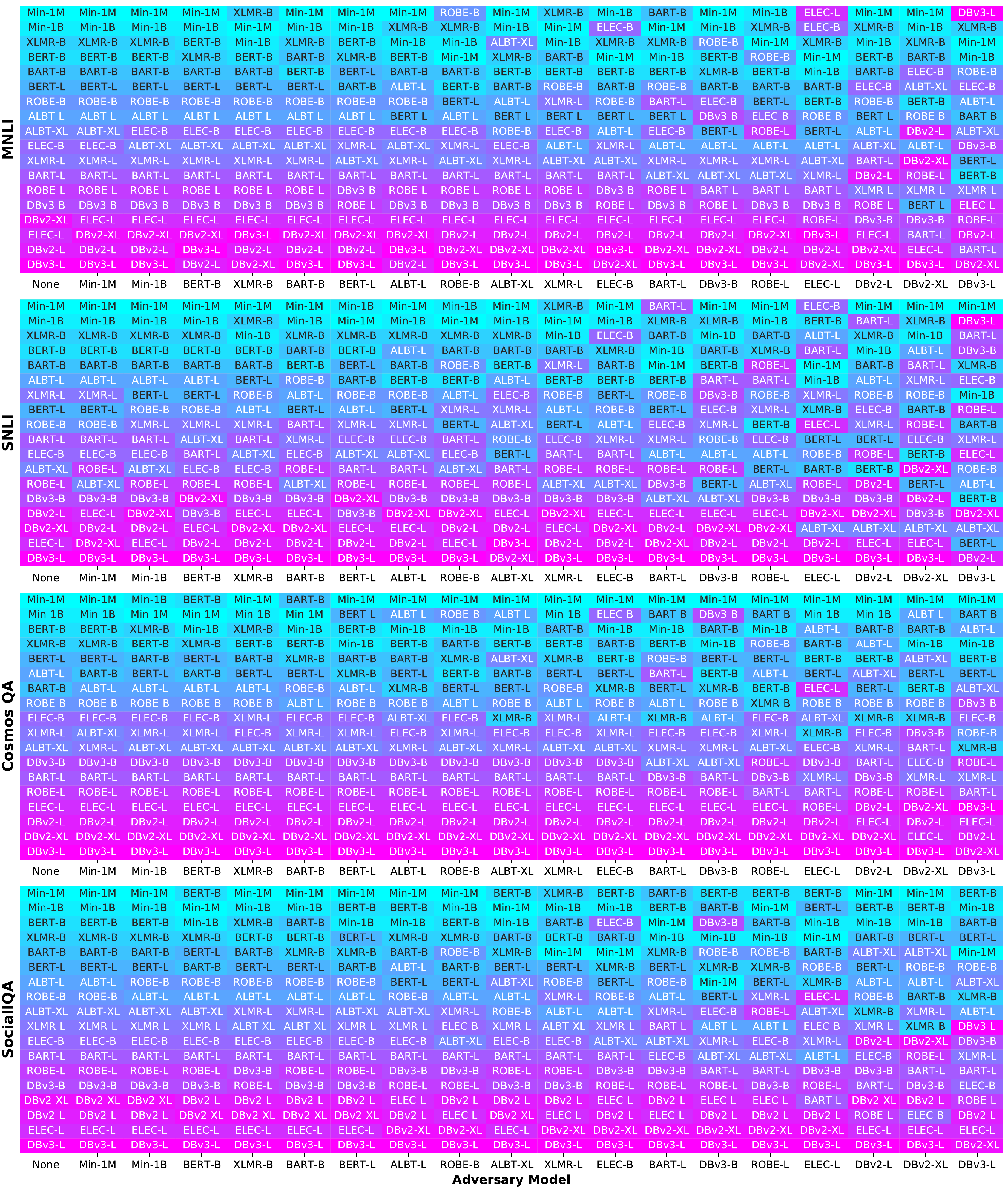}
  \caption{
    Ranked performance of fine-tuned models on validation sets filtered via AFLite using adversary models.
    For each AF Selected dataset, we sort models by their performance (Figure~\ref{fig:all_scores_curves}) from worst (top) to best (bottom).
    `None' indicates the full validation set with no filtering applied.
    We find that the sorting order of model performance is not consistent across adversary models.
  }
  \label{fig:af_rank}
\end{figure*}

\subsection{Results on AFLite Across Adversary and Fine-tuned Models}

Figure~\ref{fig:all_scores_curves} shows the results of fine-tuned models on validation sets filtered via AFLite using different adversary models. 
We present the same information in heatmaps in Figure~\ref{fig:all_scores} in the Appendix.
We emphasize that the fine-tuned models that we evaluate are trained entirely separately from the partially tuned models used to learn representations $\Phi(X)$ for the AFLite procedure.

Overall, we observe that using AFLite on successively stronger models leads to lower performance across all fine-tuned models, across all four tasks.
For MNLI and SNLI, using a sufficiently strong adversary model for filtering is sufficient to push the performance of all tuned models to only slightly above chance: For instance, while most models score between 80-90\% on the unfiltered MNLI validation set, filtering using AFLite with DeBERTa\textsubscript{RTD}-Large results in no model scoring better than 45\%.
For Cosmos QA and SocialIQA, the impact of adversarial filtering is not as large, with most models still scoring far above chance even when filtering with the strongest models.
We speculate that this is because the two multiple-choice datasets are more challenging for our models than the NLI tasks, with scores not yet saturating to the same degree, so the weak classifiers in the AFLite algorithm may not as easily filter out the easy examples.
However, we still observe the same consistent trend that filtering with stronger models leads to lower performance across the board.

We also observe a mild pattern of the weakest models (MiniBERTas, and BERT-Base) performing slightly better as stronger adversaries are used in MNLI, SNLI, and SocialIQA.
One explanation that there are certain evaluation examples that are more likely to be correctly predicted and filtered out by weaker models than stronger models.
Another possibility is that weaker models rely on easily learned heuristics \citep{mccoy-etal-2019-right}, and hence the weak classifiers in AFLite consistently select examples that go against these heuristics, which weaker models also subsequently perform worse than chance on.
In contrast, stronger adversaries might correctly label these examples, and hence they get filtered out.

\subsubsection{Impact on Model Comparison}

Evaluation datasets are often used to compare models, so we pay particular attention to the impact of adversarial filtering on the resulting relative order of model performance.
For each adversary model, we evaluate each fine-tuned model on the resulting filtered dataset and sort the models by their performance. 
We show the sorting order of models in Figure~\ref{fig:af_rank}.
We find that the order of model performance is generally not consistent across adversary models.
This is the case even if we ignore cases where the fine-tuned model shares the same pretrained model as the adversary model, which we address below.
For MNLI and SNLI, evaluating on the datasets filtered by stronger adversaries appears to greatly distort the relative performance of models.
For Cosmos QA and SocialIQA, we observe the trend that even when filtering with stronger adversaries, stronger models (based on performance on the unfiltered datasets) still tend to rank better than weaker models, but the ranking order is still not consistent across adversaries.

One possible interpretation of this result is that adversarial filtering may not give us evaluation data that is reliable for benchmarking and comparing models.
An alternative interpretation is that as stronger adversary models are used, a larger proportion of remaining examples are challenging and therefore models are more likely to perform at chance on them.
As such, we ought to expect stronger adversaries will lead to more noise in performance and randomness in the model rankings.
In the extreme, using the strongest adversary model, if the weak classifiers in the AFLite are just as capable as the fine-tuned model, all models should be performing at chance on the remaining examples.
While performance against the strongest adversarially filtered datasets is still above chance for most models, we see that in MNLI and SNLI, all models appear to converge to a much smaller range of performance (35\%--45\%), meaning that a small variation in the number of correctly predicted examples can lead to a large change in model rank.
This can contribute to a distorted ranking of models.

% We might also be concerned that the impact of adversarial filtering on performance might be disproportionately large if the fine-tuned model and adversary model are based on the same pretrained model. 
% We find that this is indeed the case.
% In Figure~\ref{fig:self_impact_rank}, for each fine-tuned model, we rank the adversary models based on how much they negatively impact performance and show the rank of using the same adversary model as the fine-tuned model.
% We observe that for the stronger half of the models, the model is almost always its own the first or second best adversary.
% For instance, RoBERTa-base is its own strongest adversary on MNLI and Cosmos QA, even though many models perform than it on the unfiltered dataset. 
% For the weaker models this is not the case, as the stronger models are often better adversaries than the model itself.
% Hence, we highlight that using a model itself to evaluate the efficacy of an adversarially filtered dataset, even when tuning on different seeds, is not a good measure of the difficulty the resulting dataset.

\begin{figure}[t]
\centering
    \includegraphics[width=0.65\linewidth]{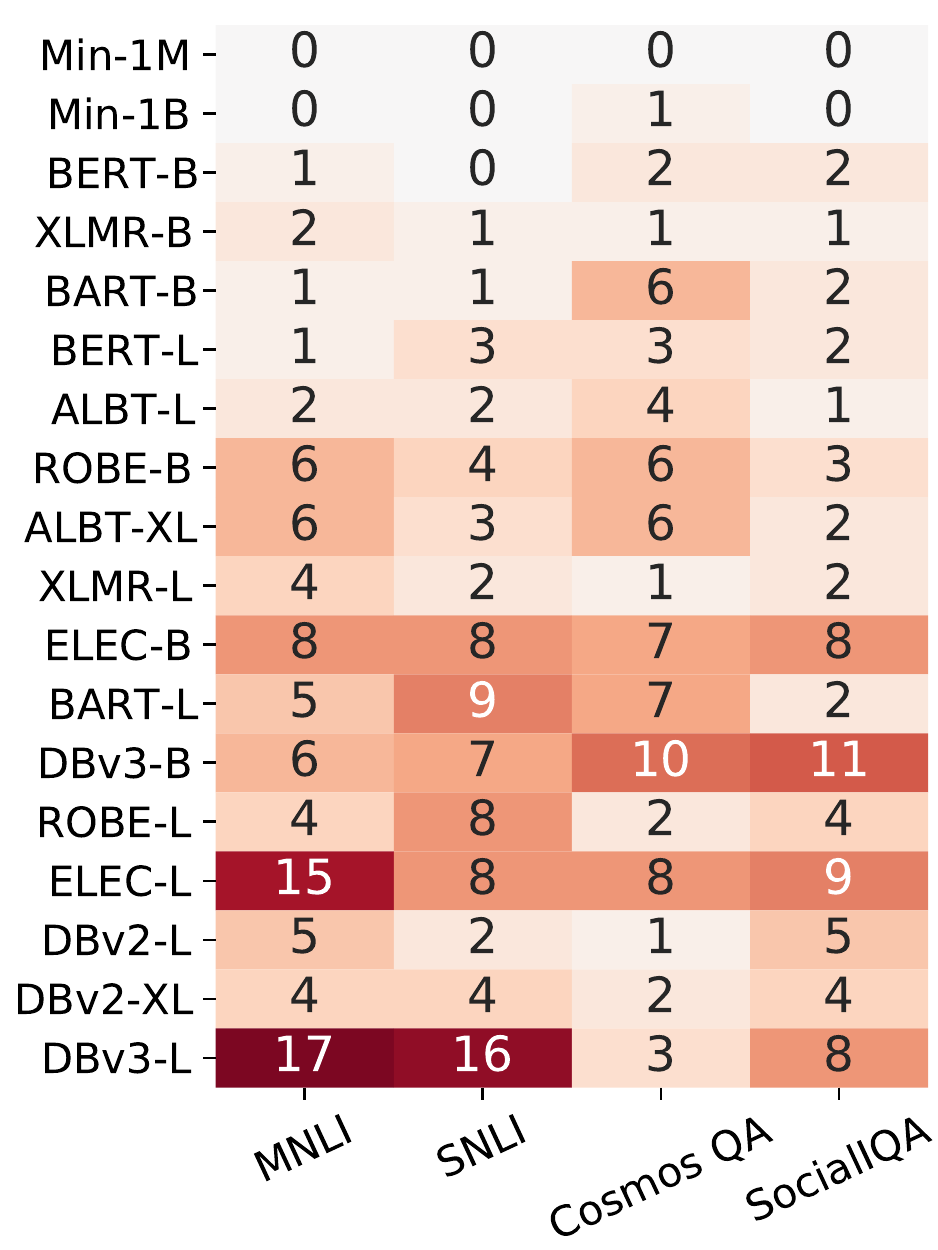}
  \caption{
    For each fine-tuned model, we compute the change in rank (1=best, 18=worst) from evaluating on the full evaluation set, and on the dataset filtered using the same pretrained model for the adversary.
    In almost all cases, filtering on the same pretrained model leads to a fall in ranking, indicating that the model is disproportionately affected by filtering with itself.
  }
  \label{fig:rank_change_from_self_filtering}
\end{figure}

We might also be concerned that the impact of adversarial filtering on performance might be disproportionately large if the fine-tuned model and adversary model are based on the same pretrained model.
To measure this, we compute the rank of each model when no filtering is applied, and show how much the rank changes when filtering using the same pretrained model as the adversary.
Ideally, if there is no model-specific bias to the filtering, there should be no change in model ranking, at the difference should be zero.
However, as we show in Figure~\ref{fig:rank_change_from_self_filtering}, the impact of filtering with the same pretrained model is disproportionately large, with all models except the weakest ones---which by definition cannot fall in rank---falling several positions in relative rankings.
There is no case where the rank improves when filtering with the same pretrained model.
This implies that adversarial filtering for evaluation sets can be very sensitive to the choice of model, and the resulting dataset can be unfairly challenging if the adversary and evaluated models are based on the same pretrained model.

\subsection{Label Agreement}

\begin{figure}[t]
    \includegraphics[width=1.0\linewidth]{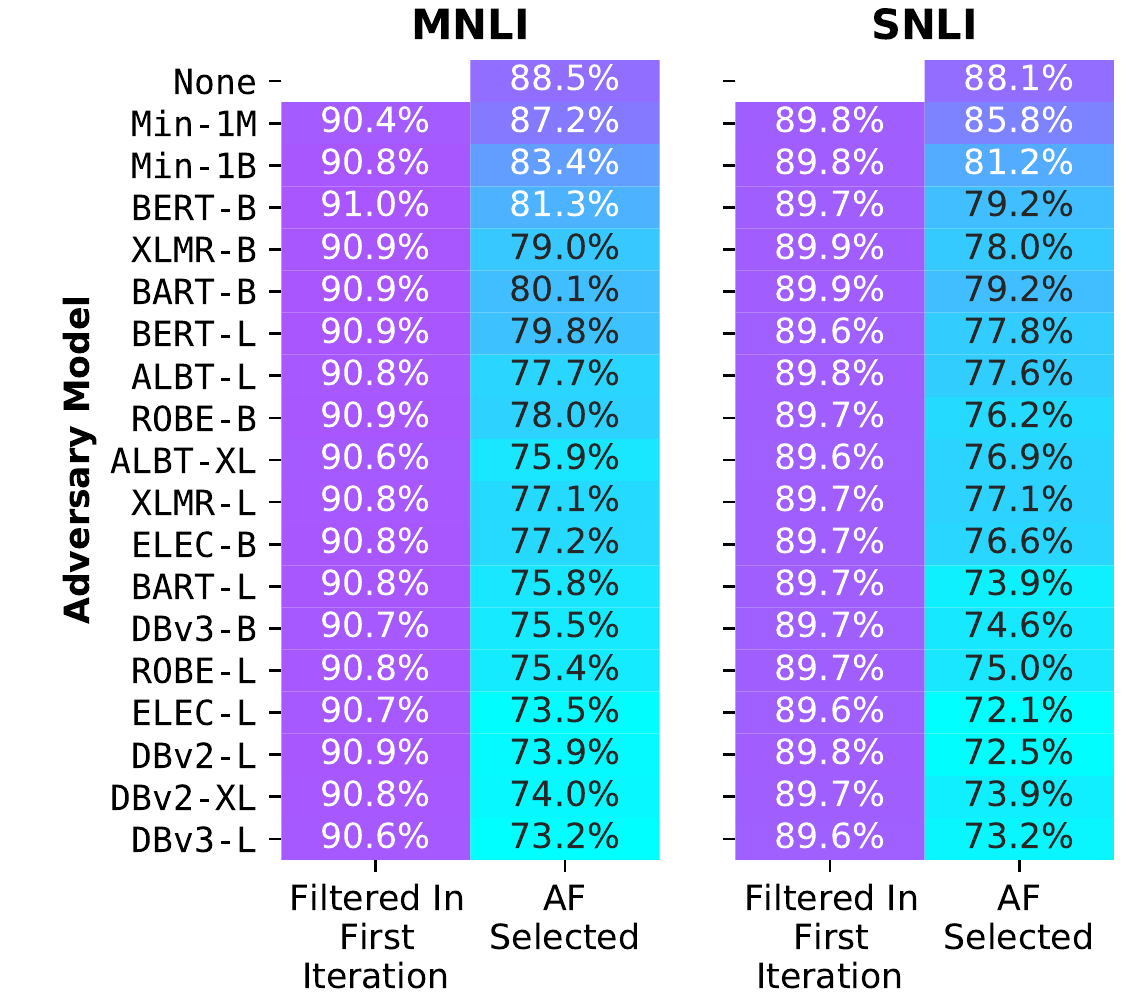}
  \caption{
    Label agreement among the adversarially filtered datasets from human annotators. 
    \textit{AF Selected} indicates examples that are not filtered out.
    \textit{None} indicates no filtering applied i.e. the agreement over the full validation set.
    Label agreement is very high for first-pass filtered examples for all models.
    On the other hand, label agreement for the remainder datasets falls as better adversary models are used, indicating that AFLite may be selecting for the examples with the most ambiguity or labeling noise.
  }
  \label{fig:agreement}
\end{figure}

To investigate the kinds of examples being identified as challenging via AFLite, we make use of the per-annotator labels provided with the MNLI and SNLI datasets.
In the original data creation procedure, each validation-set example is annotated by 5 crowdworkers, and candidate examples are only accepted if at least 3 out of 5 crowdworkers agree on the label, which is then used as the example's gold label.
We show in Figure~\ref{fig:agreement} the average rate of annotator agreement in the AFLite-selected validation sets across adversary models.
% 100\% indicates complete agreement while 60\% indicates only 3 out of 5 annotators agree on the label, which is minimum possible agreement rate.
For comparison, we also show the agreement rate among the first-pass filtered examples: examples eliminated in the very first round of the AFLite procedure.

We observe a clear pattern across both datasets that filtering with stronger adversary models selects for examples with lower annotator agreement. 
Combined with our results above on lower model performance on filtered datasets, we take this as good evidence that the AFLite procedure indeed selects for the most challenging examples.
It is unclear if these examples are challenging because they are genuinely difficult, where it is easy even for careless humans to make mistakes on them, genuinely ambiguous, or simply mislabeled.
We also note that this almost monotonic trend of increasingly selecting low-agreement examples occurs despite the fact that the numbers of examples selected via AFLite does not monotonically decrease with the strength of the adversary model.
% Figure~\ref{fig:agreement_full}
Conversely, we see that the first-pass filtered examples have consistently high annotator agreement, and that this rate does not vary across strength of the adversary models.

Oversampling low-agreement examples is not necessarily a bad thing, if they are evaluated appropriately.
\citet{pavlick-kwiatkowski-2019-inherent} and \citet{nie-etal-2020-learn} show that there can be genuine disagreement between anotators over the labels of certain examples, and argue that we should go beyond optimizing for model accuracy and instead train model to predict the full distribution of human judgements.
As easy examples seem to be highly correlated with high annotator agreement, one potential approach to construct a more challenging and discriminative benchmark could be to identify low-agreement examples, acquire additional annotations, and train and evaluate models on predicting the distribution of human labels.
However, the current format of scoring models on simple accuracy is an inadequate method of evaluating on these low-agreement examples, as the distribution of labels is reduced to a single label based on a majority vote.
Hence, if AFLite is selecting for low-agreement examples, the evaluation format should adjust according to accommodate the annotator disagreement over labels.

\section{Model-in-the-Loop Adversarially Collected Datasets}

% In model-in-the-loop adversarial data collection, human crowdworkers are tasked with writing examples that a given adversary model will incorrectly label.
% Despite the different setup, the same concern about adversarial dataset creation applies to the model-in-the-loop setup: The resulting dataset may be challenging only for the adversary model, but fails to \textit{generalize} as a challenging dataset.
% For example, this could occur if crowdworkers discover and are able to exploit a quirk of the adversary model to write examples that stump only the adversary model while being easy for other models.

In model-in-the-loop adversarial data collection, human crowdworkers are tasked with writing examples that a given adversary model will incorrectly label.
We consider two established model-in-the-loop adversarially collected datasets. 
ANLI \citep{nie2020anli}, is an adversarially written NLI dataset, collected through three iterative rounds, where the data for each round is written to be adversarial to models trained on data from previous rounds. 
BERT-Large is used as the adversary model for round 1 of data collection, while RoBERTA-Large is used for rounds 2 and 3.
Each adversary model is fine-tuned from scratch on a combination of MNLI, SNLI, and ANLI data up till that round.
AdversarialQA \citep{bartolo2020beat}, is a question-answering dataset in the format of SQuAD 1.1 \citep{rajpurkur2016squad}. 
Unlike ANLI, it consists of separately collected examples based on three adversary models: BiDAF \citep{seo2017bidaf}, BERT-Large, and RoBERTa-Large.

While both datasets come with training, validation and test data splits, we are interested in the effect of adversarial data creation on evaluation, and hence we focus our analysis on the validation splits.
For both datasets, we fine-tune models on the conventional training data for each task, before evaluating on both the standard and adversarial validation datasets.
For ANLI, we use a concatenation of MNLI and SNLI data, whereas for AdversarialQA, we use SQuAD 1.1.

\begin{figure*}[ht]
  \setlength\tabcolsep{0pt}
  \begin{tabular}{cc}
    \subcaptionbox{ANLI}{
      \includegraphics[width=.5\linewidth]{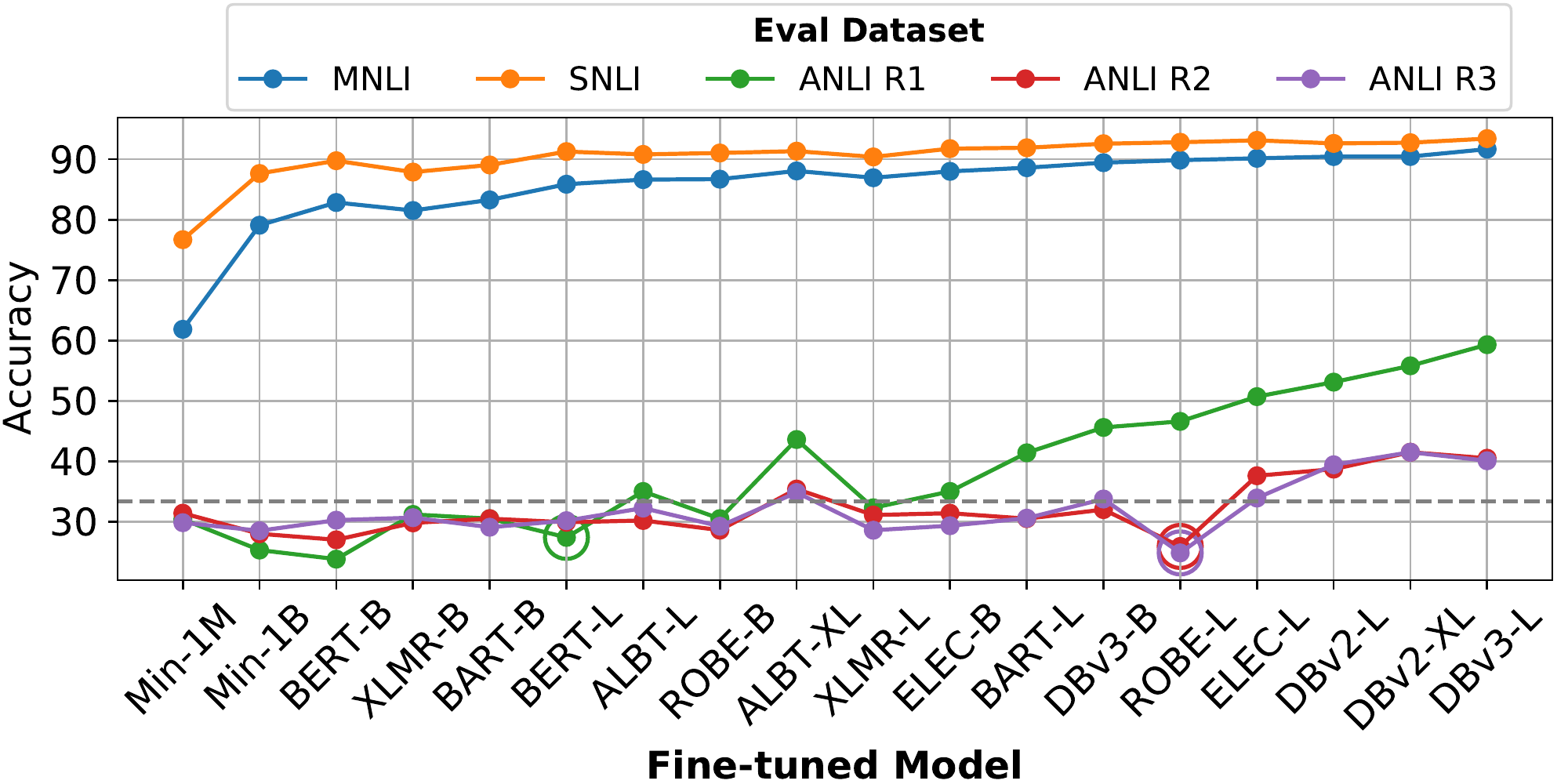}
    } &
    \subcaptionbox{AdversarialQA}{
      \includegraphics[width=.5\linewidth]{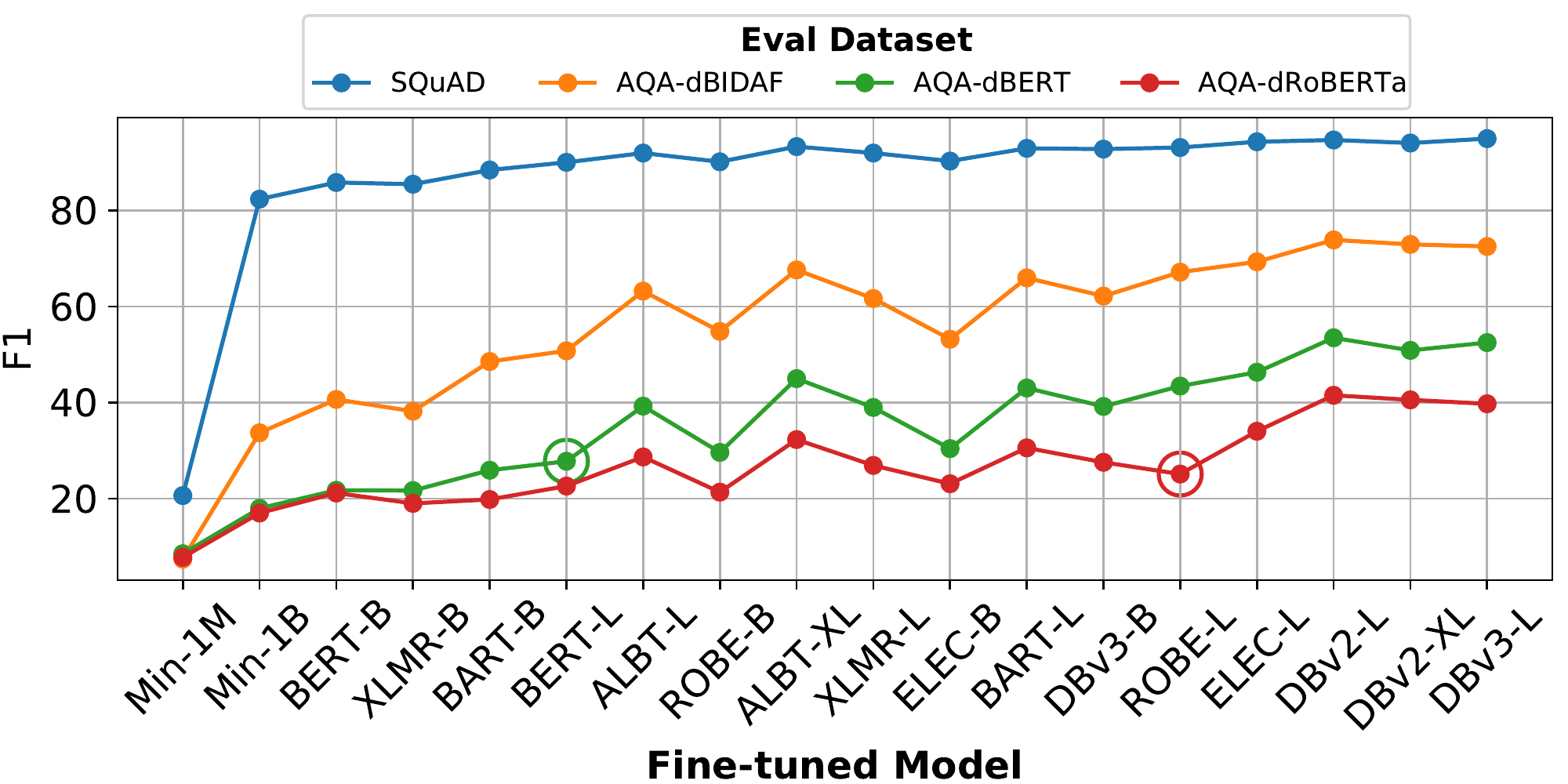}
    }
  \end{tabular}
  \caption{
    Measuring the performance of models on adversarially collected datasets.
    Exact Match scores for AdversarialQA are shown in Figure~\ref{fig:adv_datasets_aqa_em} in the Appendix.
    ANLI models are fine-tuned on SNLI and MNLI data, while AdversarialQA models are fine-tuned on SQuAD 1.1.
    For each adversarially created dataset, the corresponding base adversary model used in model-in-the-loop data creation is circled in the corresponding color for that dataset.
    Performance at chance on ANLI is shown with a dotted line.
    While adversarial dataset creation appears to create datasets that are slightly harder for the adversary model compared to other models, the resulting datasets are harder across the board for all models, with stronger models still performing relatively better.
  }
  \label{fig:adv_datasets}
\end{figure*}

We show in Figure~\ref{fig:adv_datasets} results on both model-in-the-loop datasets.
For each adversarially created dataset, we circle data points where the fine-tuned model is the same as the adversary model used in data collection.
For ANLI, we see that about half of the models perform at chance for ANLI~R1, whereas the stronger models perform significantly above chance.
On the other hand, for ANLI~R2 and R3, most models perform at chance except for the largest DeBERTa models.
Jointly, these show that the ANLI data-generating procedure leads to examples that are more difficult across all models.
However, we also observe that for ANLI~R2 and R3, the performance of the adversary model, RoBERTa-large, is markedly below chance.
This supports our observation above that while adversarial dataset creation can lower performance and raise difficulty across the board, it still tends to hurt the adversary model more than others.

We see broadly similar results for AdversarialQA.
Unlike for ANLI, models do significantly better than chance on the adversarial datasets, with almost all models obtaining above 20 F1 and 10 EM scores.
Models also generally perform poorer as the datasets are generated with stronger adversaries, and the impact is seen across all models.

% Overall, we see this as positive evidence that model-in-the-loop adversarial dataset creation is effective in creating more challenging evaluation datasets that generalize beyond the adversary model.
Compared to our more extensive experiments on adversarial filtering, there are much fewer datasets collected using different adversary models, given the financial cost and manual writing needed to obtain examples.
Hence we are unable to draw strong conclusions about the efficacy of adversarial data collection for evaluation data from the current set of results.
In particular, the adversary models used in ANLI and AdversarialQA are not among the strongest adversary models we considered in our adversarial filtering experiments, where we saw the greatest distortion in the ranking of models.
The impact of adversarial data collection on the performance of the adversary model is also hard to conclusively determine given that only two models (BERT-Large and RoBERTa-Large) fall within our scope.
However, we do find that adversarial data collection leads to harder examples with stronger adversary models.
As more work is done on adversarially collecting datasets and benchmarks are built on these datasets \citep{kiela2021dynabench}, we recommend that researchers pay close attention to the impact of the choice of adversary model and evaluate across a range of different models.

\section{Discussion}

One limitation of this study is that most of our models are encoder-only Transformer-based models, and excluded experiments on sequence-to-sequence models such as T5 \citep{raffel2019t5} and GPT-3 \citep{brown2020gpt3}, or non-Transformer models.
% However, given that encoder-only Transformers 
However, our experiments cover a diverse and comprehensive set of the prominently used models in the literature, covering a wide range of sizes (45M to 1.5B parameters), pretraining objectives (masked language modeling, sequence denoising and replaced token detection) and training corpora.
Many of these models have dominated benchmarking leaderboards and achieved state-of-the-art performance across a variety of tasks, making this is still a highly relevant sample of models to study.

We also highlight that this work has not investigated the nature of the adversarial examples outside of the impact on model performance and annotator agreement. 
Works such as \citet{williams2020anlizing} will be important for understanding exactly what examples are considered adversarial and why they are challenging to different models.

While our adversarial filtering experiments were performed on single adversary models, a possible alternative is to ensemble a diverse set of adversary models during the AFLite algorithm, or weight examples based on the AFLite example selection based on each adversary. This approach may help reduce the issue of disproportionate impact on any given adversary model's performance, and weighting evaluation across different example subsets may also potentially reduce the unstable ranking of models. However, this would significantly increase the cost of running the algorithm, and would not address the issue of oversampling low-agreement examples, which is consistent across all adversary models.

\section{Conclusion}

In this work, we have investigated two different approaches to adversarially constructing more challenging evaluation datasets.

Using a modified AFLite, we run extensive experiments performing adversarial filtering of evaluation examples and model evaluation across 18 different pretrained models.
% We find that adversarial filtering does identify more challenging example subsets, with stronger adversaries resulting in lower-scored evaluation sets. 
% However, when filtering with stronger adversary models, we also find that the resulting ranking of model performance can be unstable.
% As one might expect, adversarial filtering also greatly disadvantages models that use the same pretrained model as the adversary.
% More interestingly, adversarial filtering appears to select for examples that have low annotator agreement.
Our takeaways on the viability of adversarial filtering to create more challenging evaluation datasets are mixed.
On one hand, there is a disproportionately large impact on the performance of fine-tuned models based on the same pretrained model as the adversary, the resulting ranking of models is unstable across the choice of adversary model, especially as stronger adversaries are used, and the filtering selects for examples with low annotator agreement over labels.
On the other hand, the resulting datasets are indeed more challenging, the impact on model rankings is somewhat expected as a higher proportion of difficult examples remain after filtering, and low-agreement examples can be valuable \textit{if} an appropriate evaluation format is used that takes into account the distribution of the labels.

On our smaller set of experiments on adversarially collected datasets, we draw a set of similar conclusions.
Adversarial data collection leads to more challenging datasets, there are signs of disproportionate impact on the adversary model.

As the cost of using models goes down and their capabilities improve, we are likely to see more involvement of models in dataset creation in the future. 
Models may be used adversarially as discussed above, or used to assist in writing examples via text generation models, or used in others ways, such as automatically identifying outliers or low-quality human-written examples. 
In any of these cases, it is possible to create an adverse and undesirable feedback loop in the data creation procedure.

While we believe that adversarially constructing datasets can be a viable approach to creating more challenging evaluation benchmarks, we should take extra care to avoid the pitfalls of these approaches.
Importantly, adversarial datasets must still accurately reflect the core task or capability being measured, ideally with a diverse set of examples that have good coverage of the linguistic phenomena associated with the task.
For now, we recommend that researchers evaluate against a wide range of models where possible, and avoid measuring the difficulty of adversarial datasets using the adversary models themselves.

\section*{Acknowledgements}

We thank Vishakh Padmakumar, Naomi Saphra, Richard Pang and Nitish Joshi for their helpful comments.
This project has benefited from financial support to SB by Eric and Wendy Schmidt (made by recommendation of the Schmidt Futures program), Samsung Research (under the project \textit{Improving Deep Learning using Latent Structure}), Apple, and Intuit, and from in-kind support by the NYU High-Performance Computing Center. This material is based upon work supported by the National Science Foundation under Grant Nos. 1922658 and 2046556. Any opinions, findings, and conclusions or recommendations expressed in this material are those of the author(s) and do not necessarily reflect the views of the National Science Foundation. 

\bibliography{anthology,custom}
\bibliographystyle{acl_natbib}

\clearpage
\appendix

\section{Additional Results}

\begin{figure*}[ht!]
    \includegraphics[width=\linewidth]{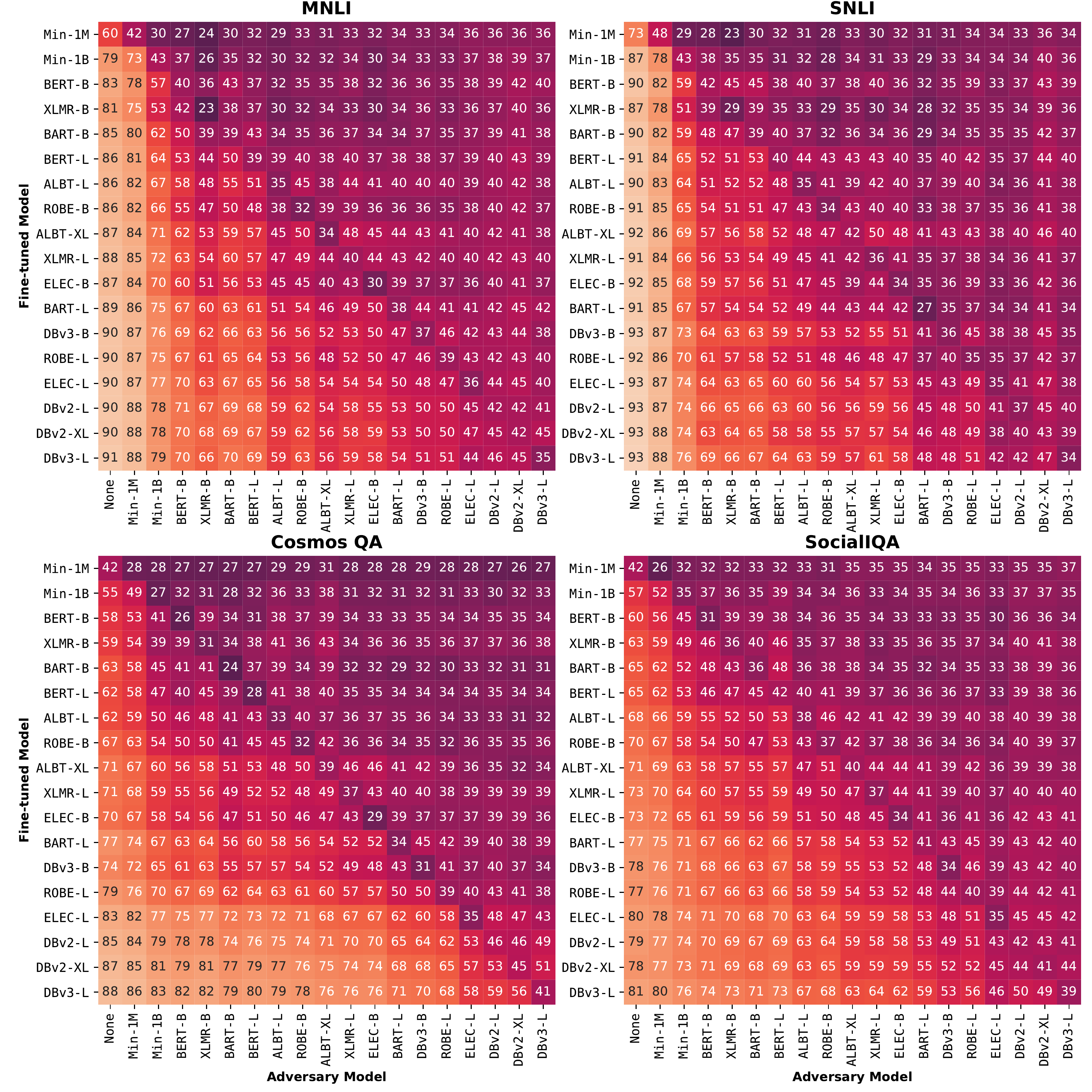}
  \caption{
    Performance of fine-tuned models on validation sets filtered via AFLite using adversary models.
    `None' indicates the full validation set with no filtering applied.
    Filtering with stronger adversary models leads to lower performance on the filtered dataset, across all fine-tuned models.
    However, filtering also tend to hurt the adversary model itself more than other models on average (darker cells on the diagonal).
  }
  \label{fig:all_scores}
\end{figure*}

\begin{figure*}[ht!]
    \includegraphics[width=\linewidth]{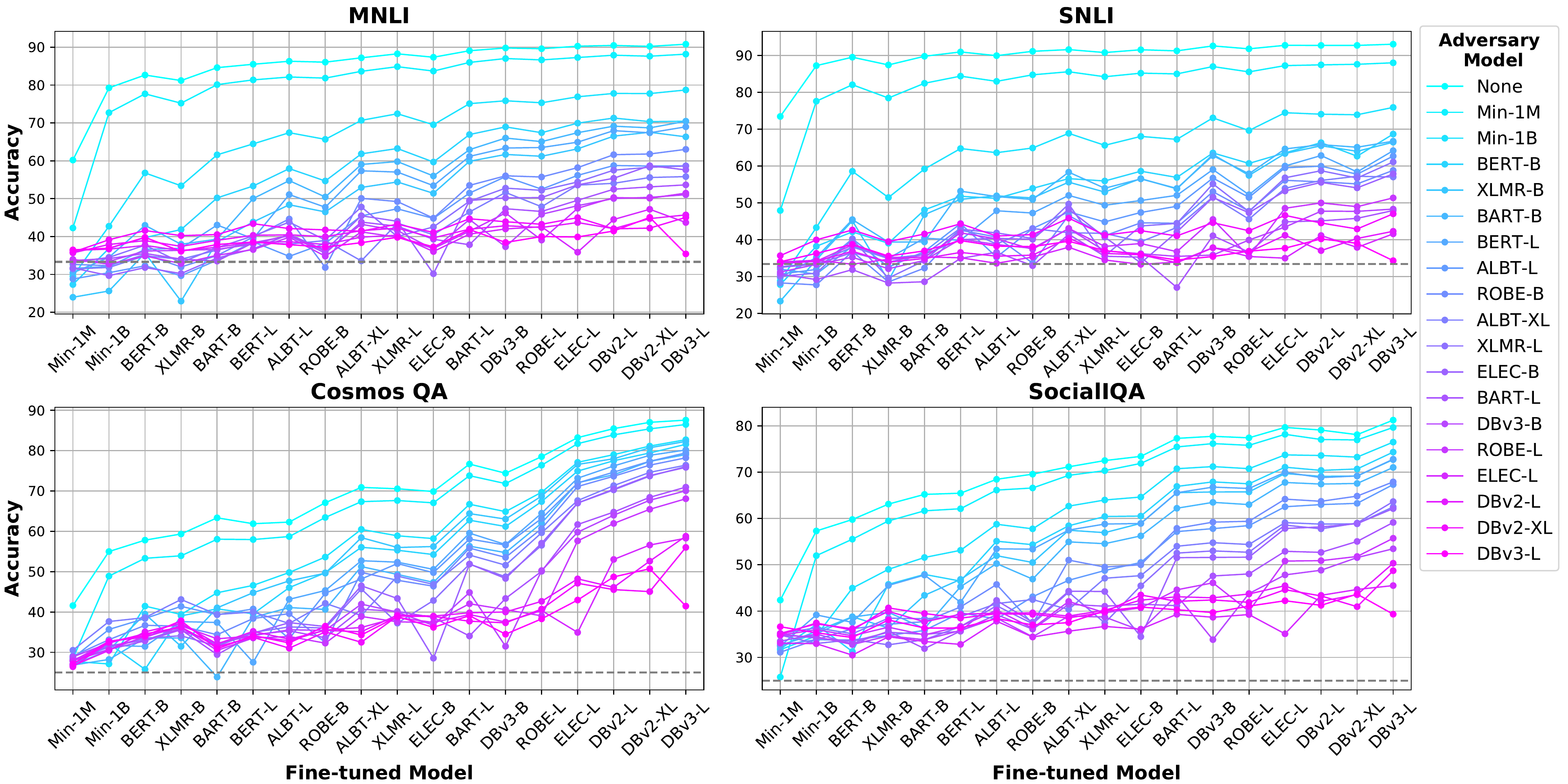}
  \caption{
    Performance of fine-tuned models on validation sets filtered via AFLite using adversary models.
    `None' indicates the full validation set with no filtering applied.
    The dotted line indicates performance at chance for each task.
    Filtering with stronger adversary models leads to lower performance on the filtered dataset, across all fine-tuned models.
    % However, filtering also tend to hurt the adversary model itself more than other models on average (darker cells on the diagonal).
  }
  \label{fig:all_scores_curves_transposed}
\end{figure*}

\begin{figure*}[t]
\centering
    \includegraphics[width=0.8\linewidth]{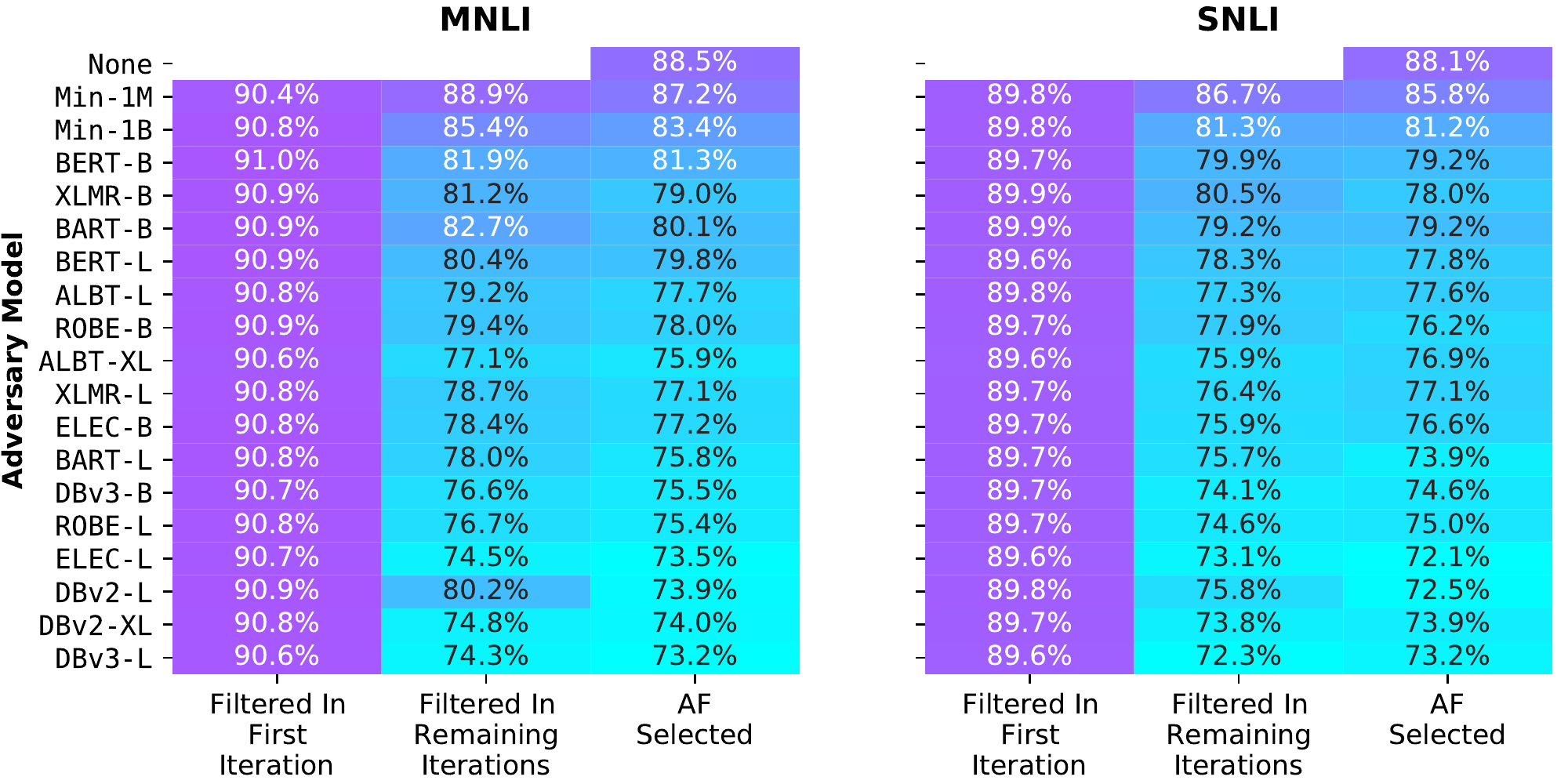}
  \caption{
    Label agreement among the adversarially filtered datasets from human annotators. 
    \textit{AF Selected} indicates examples that are not filtered out.
    Label agreement is very high for first pass filtered examples for all models.
    On the other hand, label agreement for the remainder datasets falls as better adversary models are used, indicating that AFLite may be selecting for the examples with the most ambiguity or labeling noise.
  }
  \label{fig:agreement_full}
\end{figure*}

\begin{figure*}[ht]
  \centering
  \includegraphics[width=.6\linewidth]{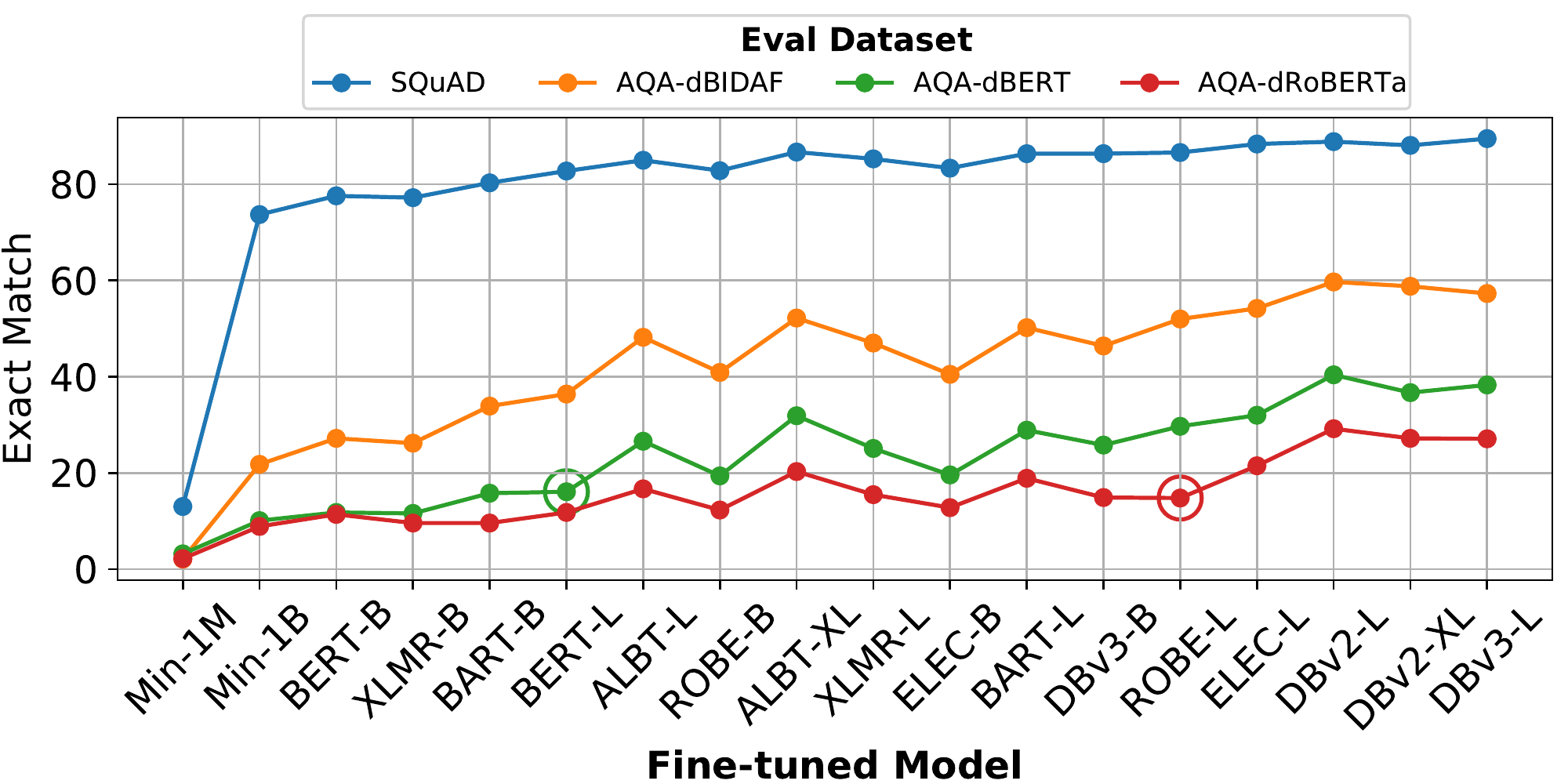}
  \caption{
    Measuring the performance of models on AdversarialQA.
    AdversarialQA models are fine-tuned on SQuAD 1.1.
    For each adversarially created dataset, the corresponding base adversary model used in model-in-the-loop data creation is circled in the corresponding color for that dataset.
  }
  \label{fig:adv_datasets_aqa_em}
\end{figure*}

Figure~\ref{fig:all_scores_curves_transposed} shows the same information as Figure~\ref{fig:all_scores_curves}, with fine-tuned models on the X-axis and adversary models shown in different curves.
Figure~\ref{fig:all_scores} shows the same information in a heatmap.
Figure~\ref{fig:agreement_full} shows the average agreement across adversarially filtered datasets, including the agreement among subsequent iterations of AFLite.
Figure~\ref{fig:adv_datasets_aqa_em} shows exact-match scores on the AdversarialQA datasets.

\section{Models}

Table~\ref{tab:models} shows additional details for each of the pretrained models used in our experiments.

\begin{table*}[ht!]
\centering\small
\begin{tabular}{lccrc}
\toprule
    Model & Abbreviation & Reference & Parameters & Training Objective \\
    \midrule
    MiniBERTa Small 1M & Min-1M & \citet{zhang2020miniberta} & $\sim$45M & Masked language modeling \\
    MiniBERTa Base 1B & Min-1B & \citet{zhang2020miniberta} & $\sim$100M & Masked language modeling \\
    BERT-base (cased) & BERT-B & \citet{devlin-etal-2019-bert} & $\sim$100M & Masked language modeling + NSP\\
    BERT-large (cased) & BERT-L & \citet{devlin-etal-2019-bert} & $\sim$340M & Masked language modeling + NSP \\
    XLM-R-base & XLMR-B & \citet{conneau2020xlmr} & $\sim$100M & Masked language modeling \\
    XLM-R-large & XLMR-L & \citet{conneau2020xlmr} & $\sim$340M & Masked language modeling \\
    BART-base & BART-B & \citet{lewis2020bart} & $\sim$100M & Text infilling + Sentence permutation \\
    BART-large & BART-B & \citet{lewis2020bart} & $\sim$340M & Text infilling + Sentence permutation \\
    ALBERT-large (v2) & ALB-L & \citet{Lan2020ALBERT} & $\sim$18M & Masked language modeling + SOP\\
    ALBERT-xlarge (v2) & ALB-XL & \citet{Lan2020ALBERT} & $\sim$60M & Masked language modeling + SOP\\
    RoBERTa-base & RoBE-B & \citet{liu2019roberta} & $\sim$100M & Masked language modeling \\
    RoBERTa-large & RoBE-L & \citet{liu2019roberta} & $\sim$340M & Masked language modeling \\
    ELECTRA-base & ELEC-B & \citet{clark2020electra} & $\sim$100M & Replaced token detection \\
    ELECTRA-large & ELEC-L & \citet{clark2020electra} & $\sim$340M & Replaced token detection \\
    DeBERTa xlarge (v2) & DBv2-XL & \citet{he2021deberta} & $\sim$900M & Masked language modeling \\
    DeBERTa XXL (v2) & DBv2-XXL & \citet{he2021deberta} & $\sim$1.5B & Masked language modeling \\
    DeBERTa\textsubscript{RTD} Base & DBv3-B & \citet{he2021deberta} & $\sim$100M & Replaced token detection \\
    DeBERTa\textsubscript{RTD} Large & DBv3-L & \citet{he2021deberta} & $\sim$418M & Replaced token detection \\
\bottomrule
\end{tabular}
\caption{Pretrained models used in our experiments }
\label{tab:models}
\end{table*}

% \section{Fine-tuning}

\section{Hyperparameters}

Table~\ref{table:aflite_hyperparams} shows the hyperparameters for our AFLite runs.

\begin{table*}[ht!]
\centering\small
\begin{tabular}{ccccc}
\toprule
    & MNLI & SNLI & Cosmos QA & SocialIQA \\
    \midrule
$m$ & 64 & 64 & 64 & 64 \\
$t$ & 50K & 40K & 10k & 10k \\
$k$ & 10K & 10K & 500 & 500 \\
$\tau$ & 0.75 & 0.75 & 0.75 & 0.75 \\
Taken From & \citet{bras2020aflite} & \citet{bras2020aflite} & \citet{sakaguchi2020winogrande} & \citet{sakaguchi2020winogrande} \\
\bottomrule
\end{tabular}
\caption{AFLite Hyperparameters}
\label{table:aflite_hyperparams}
\end{table*}

\end{document}